\documentclass[final,5p,times]{elsarticle}

\usepackage{amssymb}
\usepackage[numbers]{natbib}

\usepackage{microtype}
\usepackage{graphicx}
\usepackage{subfigure}
\usepackage{booktabs} 
\usepackage{multirow}

\usepackage{amsmath}
\usepackage{amssymb}
\usepackage{todonotes}
\usepackage{balance}

\usepackage[binary-units=true]{siunitx}
\usepackage{booktabs}    
\usepackage{float}       
\usepackage{adjustbox}   

\usepackage{algpseudocode}
\usepackage{algorithm}
\algnewcommand{\LineComment}[1]{\State \hspace{0.5cm} \(\//\//\) #1}

\usepackage{url}
\usepackage{multirow}
\usepackage[justification=centering]{caption}

\usepackage{tikz}               			
\usepackage{xcolor}						    

\usepackage{comment}

\usepackage[utf8]{inputenc}
\usepackage[most]{tcolorbox}

\DeclareMathOperator*{\argmax}{arg\,max}  
\usepackage{graphicx}
\usepackage{pifont}
\usepackage{optidef}
\usepackage{xspace}

\usepackage{wrapfig}
\usepackage{graphicx}

\begin{document}

\begin{frontmatter}



\title{MEESO: A Multi-objective End-to-End Self-Optimized Approach for Automatically Building Deep Learning Models}


\author[]{Thanh-Phuong Pham}

\affiliation[]{organization={University of Helsinki, Department of Computer Science},
            addressline={Pietari Kalmin katu 5}, 
            city={Helsinki},
            postcode={00560}, 
            country={Finland}}




\begin{abstract}
Deep learning has been widely used in various applications from different fields such as computer vision, natural language processing, etc. However, the training models are often manually developed via many costly experiments. This manual work usually requires substantial computing resources, time, and experience. To simplify the use of deep learning and alleviate human effort, automated deep learning has emerged as a potential tool that releases the burden for both users and researchers. Generally, an automatic approach should support the diversity of model selection and the evaluation should allow users to decide upon their demands. To that end, we propose a multi-objective end-to-end self-optimized approach for constructing deep learning models automatically. Experimental results on well-known datasets such as MNIST, Fashion, and Cifar10 show that our algorithm can discover various competitive models compared with the state-of-the-art approach. In addition, our approach also introduces multi-objective trade-off solutions for both accuracy and uncertainty metrics for users to make better decisions.
\end{abstract}




\end{frontmatter}


\section{Introduction}
\label{introduction}

Deep learning~\cite{Learning-Deep:2009} (DL) has been applied in various applications such as image classification, object detection, semantic segmentation, speech recognition, etc. Nevertheless, most of the training models are manually developed by machine learning (ML) experts by running many long-time experiments that require substantial computing resources and experience. To effective and efficient use of DL, automated deep learning (AutoDL) or neural architecture search (NAS) has emerged as a potential tool that alleviates the burden for both users and researchers in designing neural networks~\cite{NASSurvey:2022, AutoDL-Survey:2022, ThomasElskenNAS:2019}. 

AutoDL or NAS is a process of automating neural network architecture design. This process automates the design of deep neural networks for a specific application. AutoDL methods have outperformed manually designed architectures in several tasks such as image classification~\cite{LearningTA:2018,Real2019AgingEF}, semantic segmentation~\cite{Chenetal:2018} or object detection~\cite{LearningTA:2018}, etc. Nevertheless, the current related work does not consider all stages of the DL pipeline such as data pre-processing, feature selection, and hyper-parameter tuning~\cite{ThomasElskenNAS:2019} at the same time. One of the main challenges for that is automatically building an optimal DL pipeline requires extremely expensive efforts in terms of time, computing resources, and human expertise.

In general, the success of DL depends on the considered training data. The data is usually updated to increase the training accuracy. Consequently, the whole pipeline needs to be improved including the retraining of the deep neural network (DNN) model with hyper-parameters tuning. Further to this, it has been shown that the DNN is usually not predictive for novel data, especially noise or error data (mislabeling, inadequate information, etc.)~\cite{Minghao:2019}. This problem poses a requirement to strengthen a DNN model by identifying the novel or strange data that do not exist in both the training and testing data sets. This requirement is achieved by considering the uncertainty in prediction of the DNN model~\cite{weightUncertainty:2015, Dropout-Bayesian-2016}.

Our goal is to tackle all the mentioned challenges for automatically building DL models. To deal with the selection of various options at each stage of a DL pipeline, such as pre-processing and hyper-parameter tuning, an end-to-end approach that discovers different solutions for the whole DL pipeline is used. To confront with the uncertainty in prediction of a model, the trade-off of a DL pipeline need to be evaluated in terms of both accuracy and the uncertainty in prediction~\cite{Dropout-Bayesian-2016}. Finally, to handle the expensive training time, a novel Bayesian approach is proposed. The aim of this novel approach is to utilize the historical executions of the training process to reduce the search space and recommend various potential neural network architectures.

As a result, we propose a multi-objective end-to-end self-optimization approach for automatically building DL models. Our approach is inspired by Bayesian Optimization~\cite{Bayesian:2015} that utilizes all existing knowledge while keeping the diversity of considered candidates. Furthermore, our approach also avoids as many as possible the actual end-to-end execution of training an architecture based on prior training executions. To that end, we propose a novel surrogate function that ranks potential architectures into ordered groups. The goal of this surrogate function is to estimate the potential architectures in advance, so the actual training executions are reduced to the potential architectures, however, the diversity of the search are still kept since the surrogate function can offer multiple architectures.

In summary, our work alleviates users' effort for applying DL to their applications and offer various trade-off deep neural network architectures to users. Experimental results on three well-known dataset MNIST~\cite{Mnist}, Fashion~\cite{Mnist-Fashion}, and Cifar10~\cite{Cifar10} show that our approach can give competitive results compared with the popular state-of-the-art Bayesian approach such as Autokeras~\cite{Auto-keras}. Our approach also introduces a multi-objective evaluation of generated DL models in terms of accuracy and uncertainty for better decisions.

The main contributions of our work are as follows:
\begin{itemize}
	
	\item Proposing a novel self-optimized approach for automatically building DL models. 
	
	\item Modeling the end-to-end DL pipeline for better selecting various options at each stage of the pipeline.
	
	\item Introducing a multi-objective evaluation for the end-to-end DL pipeline that helps users make a decision following their demands. 
	
\end{itemize}

\section{Related work}
\label{relatedwork}


The main challenge of NAS is training many deep neural networks usually requires a huge amount of time and computing resources. A lot of effort has been invested to automate the design of DNNs. In general, the variety of heuristic and meta-heuristic of AutoDL approaches can be summarized into the following categories: random search~\cite{Bergstra:2012:Randomsearch, Fabioetal2013, Wangetal2013}, grid search (or brute force)~\cite{GridSearch:2018}, reinforcement learning~\cite{NeuralAS-RL:2017, AutoMG:2022}, evolutionary algorithms~\cite{GeneticCNN:2017, NSGA-Net:2019, NSGA-Net:2020, EA:2022}, particle swarm optimization~\cite{NASSurvey:2022, AutoDL-Survey:2022, IntelliSwAS:2022}, Bayesian optimization~\cite{Auto-keras, BANANAS:2020, MOBayesian:2020}, etc.It has been shown that the searching time to discover an appropriate neural network architecture usually requires days, weeks, or even months~\cite{BANANAS:2020, Large-ScaleEvolution:2017}. Generally, AutoDL assumes that users do not have any experience with a neural network thus the searching time might take days even for small input data such as Cifar10~\cite{GeneticCNN:2017, NSGA-Net:2019}.  

Many families of meta-heuristics as evolutionary algorithms (EAs) are general approaches to deal with a search problem like NAS. Nevertheless, EAs are still a hindrance for users due to their long-time executions. To overcome this, the initial population of a NAS problem is usually a set of potential candidates that are used as the starting points for the EAs. In other words, the population initialization step is usually based on prior knowledge from hand-crafted architectures~\cite{ResNet:2016, tan2020efficientnet, IntelliSwAS:2022, EA:2022}. Furthermore, the efficiency of EAs is usually considered and improved, such as using a surrogate-assisted method~\cite{NSGA-Net:2020}. However, it is still an expensive approach because the training time of a deep neural network is naturally time-consuming, and EAs need to evaluate many candidates in each iteration with actual executions~\cite{AutoMG:2022, NSGA-Net:2020, EA:2022, NoAS-DS:2022}.

Bayesian optimization has been successfully applied in hyper-parameter tuning~\cite{Snoek2012, Martinez2014:Bayesian, HyperparameterBO:2019} and has recently emerged as a promising approach for NAS~\cite{Auto-keras, BANANAS:2020}. In contrast to other search approaches, Bayesian approaches keep track of past evaluation results and eliminate the estimation of unpromising solutions. Therefore, they are powerful tools for optimizing expensive objective functions as training a deep neural network. However, the challenge of these approaches is to construct an appropriate surrogate model that estimates the value of the objective function in advance. Different related works have their own approaches to deal with this problem~\cite{AutoDL-Survey:2022, Auto-keras, BANANAS:2020, Hutter2019AutomatedML, ReviewNAS:2022}.

It is shown that NAS space is not an Euclidean space, and it does not satisfy the assumption of the Gaussian process~\cite{Auto-keras, BANANAS:2020}. To apply the Bayesian optimization to NAS, novel approaches are required. The Autokeras~\cite{Auto-keras} introduces edit-distance neural network kernel, tree structure space, etc. BANANAS~\cite{BANANAS:2020} introduces an architecture encoding method for training neural network surrogate model. These approaches also result in more complexity for constructing a surrogate model. In addition, the kernel function that is used to generate the Gaussian distribution for NAS problem still produces many neural architectures that require actual evaluations or training executions.In this work, we introduce a novel heuristic for generating neural network architectures and a corresponding surrogate model that learns to rank the neural network architectures and reduces the performance of the searching process. 

Many related works consider the multi-objective evaluation for a neural architecture such as error rate, complexity, robustness resisting to adversarial data~\cite{Minghao:2019, NSGA-Net:2019, NSGA-Net:2020, Kotyan:2020, AutoMLSurvey:2020}, etc. In this paper, we consider a new metric for evaluating a DNN model named uncertainty in prediction. Our multi-objective evaluation consists of two metrics: accuracy and uncertainty in prediction. The uncertainty metric is an important property that is useful for recognizing the out-of-knowledge of the neural network architecture as well as recognizing the necessity of updating or retraining the architecture with novel data. With this metric, the AutoDL tool can introduce a confidence for each prediction result, so the user can make a better decision following both obtained metrics for each DNN model.

To speed up the convergence of the search, an algorithm should start with an exploration of a potential initial space that is usually based on prior knowledge~\cite{Auto-keras,NSGA-Net:2019,NSGA-Net:2020}. Since the random search space usually leads to time-consuming and inefficient search. To cope with that, our work propose the use of different heuristics based on the existing knowledge. The search space can be updated or changed during the search following considered heuristics. For example, if the considered problem is image classification, then the computational blocks that are used to construct or grow up the neural network archiectures can be Inception module~\cite{Inception:2015}, ResNet identity block~\cite{ResNet:2016}, EfficientNet~\cite{tan2020efficientnet} etc. Generally, our approach will try different additional heuristic, thus, we name our proposed method the self-optimized approach. Notably, most of the related work like Autokeras~\cite{Auto-keras}, NSGA-Net~\cite{NSGA-Net:2019,NSGA-Net:2020} introduces different approaches for generating the search space. In other words, a complete random search space does not work, therefore, an appropriate search space is necessary.

In summary, we introduce a heuristic to generate an appropriate search space for a NAS problem. We propose an end-to-end evaluation for NAS with two metrics: accuracy and uncertainty in prediction. The main contribution of our work is a novel surrogate model that reduces actual training executions. Finally, our approach repeats the searching process with different heuristics. Therefore, we name it the multi-objective end-to-end self-optimized approach. To the best of our knowledge, our work is the first work that introduces such novelties.

\section{Background}
\label{background}

\subsection{Deep Learning}
\textit{Machine learning} (ML) methods generally investigate the relation between input data and its output. This is usually learned after observing a dataset and inferring the relation between input and output values. In the field of ML, the input dataset is usually referred to as a training set or simply training data. The output can be any set of values following real observation. ML commonly has to confront a major challenge, which is to  discover the unknown function between input historical data and the corresponding output. 

\textit{Deep learning} is a subset of machine learning that focuses on building artificial neural networks to study the complex relationship between input and output. To achieve the target, deep neural networks are usually built such as deep belief networks, recurrent neural networks, convolutional neural networks, etc. Deep neural networks have been successfully used in different fields including computer vision, audio recognition, machine translation, etc. They have produced results comparable to, and in some cases surpassing human expert performance~\cite{NASSurvey:2022, LearningTA:2018, Real2019AgingEF, Chenetal:2018}.


\textit{Uncertainty in deep learning:}
There are two main different types of uncertainty in deep learning named epistemic and aleatoric uncertainty~\cite{uncertainty:2019}. The former describes what the training model does not know due to the limitation of training data. The latter describes the natural stochasticity of the training data, where the error could exist due to unintended incidents. We target to evaluate the epistemic uncertainty of a deep neural network architecture since it is an important property for real-world applications~\cite{weightUncertainty:2015, Dropout-Bayesian-2016}. For example, a model is not only accurate but also certain. In other words, the model can predict well for training data, and it can recognize novel data. If the uncertainty is too high, then users could have their appropriate decisions. 

\subsection{Multi-objective optimization}
For the sake of completeness, we formally define the multi-objective optimization problem (MOP), Pareto optimal solutions, and Pareto front. In this section, we define the minimization of all objectives, however, the following definition is general for a MOP since a maximization problem can be defined as a minimization too. Formally, the problem is defined in equation (\ref{optimization}).

\begin{equation}
	\label{optimization}
	\begin{aligned}
		& \underset{x \in \Omega}{\text{minimize}}
		& & F(x) = (f(x_1), f(x_2), \ldots, f(x_n)) \\
		& \text{subject to} & &  c_i(x) \leq 0, \forall i \in \Upsilon \\
	\end{aligned}
\end{equation}

where $f_i$ are $n (\geq 2)$ conflicting objectives that have to be simultaneously minimized and the vector $ x = ([x_1, x_2, \dots, x_n])^T $ is the vector of decision variable that belongs to a non-empty set $\Omega$ called decision space and satisfies the constraint $c_i$. $\Upsilon$ is the set of all possible constraint values of $c_i$

\textbf{Definition 1} 

A vector $ h = [h_1, h_2, \dots, h_n]^T $ dominates another vector $ k = [k_1, k_2, \dots, k_n]^T $ if and only if $\forall i \in [1, 2, \dots, n]$, $ h_i \leq k_i$ and $\exists j \in [1, 2, \dots, n]$, $h_j < k_j$. This is denoted $h \prec k$. Two solutions are non-dominated if neither of them dominates the other. 

\textbf{Definition 2} 

In a general multi-objective optimization problem, a Pareto optimal set is defined as a non-dominated set in the decision space that is formally described in equation (\ref{Pareto_optimal_set}) 
\begin{equation}
	\label{Pareto_optimal_set}    
	\begin{aligned}
		S = \{ \forall x \in \Omega |\; \nexists \, x^* \in \lambda, f(x^*) \prec f(x) \} \\
	\end{aligned}
\end{equation}
where $\lambda$ is a set of all decision variables in $\Omega$ that are feasible (satisfy all constraints).

\textbf{Definition 3} 

For a given pareto optimal set \textit{P}, the Pareto front is the value of pareto optimal set corresponding to f as it is defined by equation (\ref{Pareto_front})
\begin{equation}
	\label{Pareto_front}    
	\begin{aligned}
		PF = \{ f(x), x \in S \}
	\end{aligned}
\end{equation}
A pareto front represents the trade-off results that allow decision makers to explore the possible non-dominated solutions following their needs. In this way, users do not have to set their references in advance, conversely, they can choose the preferred regions after the Pareto front is calculated.

\subsection{Hyperparamter Optimization}
Hyperparameter optimization (HO) in ML is the process that considers the training variables set manually by users with pre-determined values before starting the training~\cite{HyperparameterBO:2019, Auto-weka:2017}. This process aims to search hyper-parameters of a given machine learning algorithm that return the best results. Formally, HO can be represented in an equation form as follows:
\begin{equation}
	\begin{aligned}
		x^{*} = \argmax(x) \in \mathcal{X} g(x)
	\end{aligned}
\end{equation}

where $g(x)$ is the objective function for minimizing such as error rate, $x^*$ is the set of hyper-parameters that give the lowest value of the error rate and $x$ is the variable that can take any value in the consider domain $\mathcal{X}$. In case the ML method has many hyper-parameters such as a deep neural network, the optimization process is extremely expensive in terms of time and computing resource consumption. 

\section{Problem Definition}
\label{problemdefinition}

\subsection{Motivation}

End-to-end optimization of a DL pipeline is important for automatically achieving optimal results. Every appropriate selection of each stage in the pipeline will contribute to the improvement of performance, accuracy, and other metrics of the whole pipeline. For instance, in the data pre-processing stage input data usually requires to be normalized and standardized before being sent  to a DNN for training. There are different situations where optimization is necessary to adjust a selected DNN, such as the diversity of the input dataset might require the construction of more and more complex deep neural network architectures. Furthermore, it is also apparent that a right architecture of a neural network has to be tuned with different hyper-parameters to become a better predictor or classifier~\cite{HyperparameterBO:2019}. 

A DNN usually has many hyper-parameters, considering the tuning of all the parameters often takes a huge amount of time. To reduce human effort in applying DL to real-world applications, an automatic approach that gradually improves the training model is necessary~\cite{AutoMLSurvey:2020}. These approaches should frequently update the model based on currently obtained results. This automation aims to incrementally enhance the prediction model upon prior training. After a good architecture is discovered, the algorithm will look for solutions around that architecture. The idea is inspired by Bayesian optimization. In addition, our work goes step by step across different heuristics with an end-to-end evaluation of a potential architecture, and we evaluate the pipeline on multiple metrics. Therefore, an end-to-end optimization has a greater search space than the pure hyper-parameter optimization of deep neural networks. 

\subsection{Formal Definition}
In this section, we generally define the optimization problem for a ML pipeline. Without loss of generality, let $x_i$ denotes a feature vector and $y_i$ is the respective target value. The training and testing datasets of a ML problem are described as follows:
\begin{align}
	D_{train} = \{(x_1,y_1),\ldots , (x_n, y_n)\} \\
	D_{test}  = \{(x_{k+1}, y_{k+1}, \ldots, (x_{k+i}, y_{k+i}))\}
\end{align}

where the values of testing dataset ($x_{k+j}$, $y_{k+j}$) are different from values of training dataset  $\forall j$  $\in$ $[1,\ldots, i]$. Let's define each stage of a pipeline as a set $S_i$. Then the general pipeline can be defined:
\begin{equation}
	\label{dl_pipeline}
	P = \{S_1, S_2, S_3, \ldots ,S_m\}
\end{equation}

where $P$ is the pipeline and $S_i$ is a stage of $P$. Each $S_i$ contains possible solutions for the corresponding stage of $P$.
Following the above defined equations, the end-to-end pipeline optimization problem can be described such as follows:
\begin{equation}
	\label{optimization_pipeline}
	\begin{aligned}
		& \underset{D_{train}, D_{test}}{\text{minimize}}
		& & F (S_1 * S_2 * S_3 * \ldots * S_m) \\
		& \text{subject to} & &  c_{t}(P) \leq 0, \forall t \in \Upsilon \\
	\end{aligned}
\end{equation}

where $P$ is the pipeline defined in the equation (\ref{dl_pipeline}) and the $c_t$ is the constraint of the problem. The $\Upsilon$ is the set of all possible constraints for the optimization problem, for instance, cost, performance, accuracy, uncertainty, etc. The operator "*" is the Cartesian product of sets. Therefore, the Cartesian product of all $S_i$ describes all possible combinations of $P$. 

\section{Multi-objective End-to-end self-optimized approach}
\label{sec:end-to-end-optimization}

\subsection{Considered Objectives}
\label{considerobjs}
Although a DNN usually works well with most of the prediction problems, most of the time it is not clear how certain the prediction is. There is an obstacle to deploying predictive models in high confidence fields such as medicine, aviation, judiciary, etc. Generally, a DNN model needs to be able to recognize novel data and evaluate the uncertainty of its prediction. This property is called epistemic uncertainty and it is an important property for delivering DNN models to the fields that need high reliability~\cite{weightUncertainty:2015, Dropout-Bayesian-2016, uncertainty:2019}. 

We propose the evaluation of the epistemic uncertainty along with the accuracy as two important metrics of a deep neural network. There are two popular methods for dealing with the uncertainty in prediction of a DNN: Variational Inference~\cite{VariationalIn:2017} and Monte Carlo Dropout~\cite{Dropout-Bayesian-2016}. The latter is used to evaluate the uncertainty of the DNN model in our work. Monte Carlo Dropout is proposed by~\cite{Dropout-Bayesian-2016} to evaluate the model uncertainty. The method is briefly explained as follows.


Let's define the training neural network model with a concrete dropout rate $m_{d_i}$, where $m$ stands for the model and the ${d_i}$ is the dropout rate respectively~\cite{Dropout-Bayesian-2016}. To derive the uncertainty of the model $m$ for one sample input $t$, we collect N prediction of $m$ on $t$ with different ${d_i}$. Next, we compute the average and the variance of this sample which are the mean of the model (posterior distribution) for this sample and the estimate of the uncertainty of the model regarding $t$ respectively. 

The predictive posterior mean of the model $m$ with a dropout rate $d_i$ is calculated:
\begin{equation}
	\label{posterior_mean}
	\begin{aligned}
		p = \frac{1}{N} \sum m_{d_i}(t)
	\end{aligned}
\end{equation}

And the uncertainty of the model $m$ regarding $t$ is described in the equation (\ref{uncertainty_estimation}):

\begin{equation}
	\label{uncertainty_estimation}
	\begin{aligned}
		u = \frac{1}{N} \sum [m_{d_1} - p]^2
	\end{aligned}
\end{equation}
Properly including uncertainty in evaluating, a DNN can also help to debug prediction results, and making the network more robust against completely new input data. As a result, we consider uncertainty and accuracy as two objectives of our optimization approach.

\subsection{Initial space}
\label{InitialSpace}
To speed up the convergence, a search algorithm usually starts with a potential initial space that is usually based on prior knowledge or heuristic. In DL, to deal with a complex problem, additional neural network layers are added to a deep neural network which usually leads to the improvement of prediction accuracy. The main idea behind adding more layers is that these layers will learn more complex features. For example, ResNet~\cite{ResNet:2016} is a specific type of neural network that can gain accuracy from considerably increased depth. However, this is not always the case where the decreased depth and the increased width of the residual network can be more efficient in some concrete situations~\cite{WideResNet:2016}. Deeper and deeper networks are prone to over-fitting, thus, wider networks are more preferred in some cases~\cite{Inception:2015}. In addition, it is shown that the balance among network depth, width and resolution can result in better accuracy and performance~\cite{tan2020efficientnet, Auto-keras}. 

Following the above observations, a potential initial search space should consist of balanced network architectures. In this work, we initialize the search space of neural network architectures based on some well-known neural network unit blocks. The search space is updated or changed during the search process following some considered heuristics. For example, if the considered problem is image classification, then the computational blocks that are used to construct or grow up the neural network can be Inception module~\cite{Inception:2015}, ResNet identity block~\cite{ResNet:2016}, etc. The network architectures will grow up wider and deeper in a balanced manner. In other words, both of the dimensions are scaled together instead of one. 

\begin{algorithm}[t!]
	\caption{ Surrogate Function - Learn to Rank }
	\hspace*{\algorithmicindent} \textbf{Input:}                                \\
	\hspace*{\algorithmicindent} \textit{\hspace{1cm} candidates (arches)}      \\
	\hspace*{\algorithmicindent} \textit{\hspace{1cm} group of ranking (groups)}\\
	\hspace*{\algorithmicindent} \textit{\hspace{1cm} metric    (m)}   \\

	\label{alg:surrogate}
	\begin{algorithmic}[1]
		\Procedure{$surrogate\_model$}{$ arches, groups, m $}   
		\\
		\LineComment{read architecture information}
		\State $ data \leftarrow read\_database(arches) $
		
		\\
		\LineComment{create groups for ranking}
		\State $ data \leftarrow group\_generation(groups) $
		
		\\
		\LineComment{split data for training model} 
		\State {$x\_train, y\_train \leftarrow generate\_train(data)$}
		
		\\
		\LineComment{train surrogate model} 
		\State $ surro\_model \leftarrow Ranker(x\_train, y\_train, m) $
		
		\\
		\LineComment{return surrogate model} 
		\State $ return \hspace{0.5cm} surro\_model $
		
		\EndProcedure
		
	\end{algorithmic}
\end{algorithm}

\subsection{Surrogate function}
\label{sec:surrogate_func}
The surrogate function is used to approximate the actual function, for instance, training a neural network. This function usually helps to estimate a large range of candidates and eliminate less important ones in the search space. It is usually efficient to have a surrogate function when the actual function is extremely expensive in terms of time, and computational resources. 

To be more effective concerning the convergence of the algorithm, the surrogate function should be re-calculated to incrementally learn from the historical runs. The longer the algorithm runs, the closer the surrogate function comes to resembling the actual function. From the estimated results, one or more candidates will be selected and evaluated with the actual function.

It is a challenging task to construct a suitable surrogate function for the NAS problem since it requires an efficient method for comparing the neural networks. Jin et. al~\cite{Auto-keras} propose an edit-distance kernel for calculating the distance between neural networks. White et.al~\cite{BANANAS:2020} introduce an encoding for neural network and build a meta neural network for comparing the encoding vector instead of directly comparing the neural networks. 

In this paper, we present a novel approach  to construct a surrogate model that can rank the neural network architectures into different groups. This idea comes from the observation presented in section \ref{InitialSpace}, where the wider, deeper, and balance architectures might have different effective results in various cases. The details for this surrogate function are described by the algorithm~\ref{alg:surrogate}.

Generally, we assume that there are some historical runs including various architectures. The obtained historical data are assigned the number of groups line (6-7). Then the data is split into training and testing data for training the surrogate model. In this work, we use the \textit{LightGBM}~\cite{LightGBM:2017} algorithm as the $Ranker$ (line 13) for training this surrogate model. The main goal of this surrogate model is to reduce the search to the ranking groups, thus, it helps to reduce also the execution time of the optimization process.

\subsection{Acquisition function}

To navigate the search in response to the surrogate function, it is necessary to have a function that is used to interpret and score the response from the surrogate function. This type of function is usually called the acquisition function. To score the results from the surrogate function, the acquisition function has to consider all objectives of the problem. The function is responsible for scoring or estimating the value of a given candidate (as input) is worth evaluating with the actual function. After having the scores of the acquisition function, the optimal candidates will be selected. The selected candidates will be evaluated via the actual function.

\begin{algorithm}[t!]
	\caption{ Acquisition Function - Optimizing Search }
	\hspace*{\algorithmicindent} \textbf{Input:}                                   \\
	\hspace*{\algorithmicindent} \textit{\hspace{1cm} surrogate      (model)}      \\
	\hspace*{\algorithmicindent} \textit{\hspace{1cm} search\_space  (space) }     \\
	\hspace*{\algorithmicindent} \textit{\hspace{1cm} number\_arches (arches)}     \\
	\label{alg:acquisition}
	\begin{algorithmic}[1]
		\Procedure{$acq\_func$}{$ model, space, arches $}     
		\\
		\LineComment{read search space}
		\State $ data \leftarrow search\_space(space) $
		
		\\
		\LineComment{extract architectures}
		\State $ data \leftarrow extract\_architectures(space) $
		
		\\
		\LineComment{find potential architectures} 
		\State {$po\_arches \leftarrow model(data) $}
		
		\\
		\LineComment{sort the results following rank order} 
		\State {$po\_arches \leftarrow sort\_values(po\_arches)$}
		
		\\
		\LineComment{return $arches$ architectures} 
		\State $ return \hspace{0.5cm} po\_arches(arches) $
		\EndProcedure
		
	\end{algorithmic}
\end{algorithm}

In multi-objective Bayesian optimization, it is assumed that the objectives are mutually independent in the objective space. Each objective is approximated by a surrogate model. The goal is to compute a set of trade-off solutions according to the considered objectives. To this end, we target to creating several solutions at the same time instead of taking only one solution. 

In other words, instead of taking one architecture based on its current accuracy, we will take multiple ones that might lead to a trade-off between considered objectives (accuracy and uncertainty). Our acquisition function allows setting the desired number of solutions $arches$ that will be taken after ranking by the surrogate model. Consequently, the function will return a set of $arches$ architectures instead of one, line $16$ in the algorithm~\ref{alg:acquisition}.

\subsection{Proposed algorithm}
\label{sec:proposed-algorithm}
\textbf{End-to-end:} Many important stages affect the final results of a model such as data pre-processing, compile and train the architectures, etc. In this work, we illustrate the general end-to-end evaluation for each neural network architecture in the algorithm~\ref{alg:endtoend}. Concretely, after the data is loaded, it is usually necessary to have a pre-processing step before the data can be used for training the neural network model. Furthermore, the training step often requires hyper-parameters tuning~\cite{Auto-weka:2017, HyperparameterBO:2019} to achieve expected results such as high accuracy or low error rate. As a result, when a potential architecture cannot provide the expected accurate prediction, each stage of the end-to-end pipeline should be reconsidered carefully.

\begin{algorithm}[t!]
	\caption{ End-to-End Evaluation for Architectures }
	\hspace*{\algorithmicindent} \textbf{Input:}                                        \\
	\hspace*{\algorithmicindent} \textit{\hspace{1cm} search\_space  (space) }          \\
	\hspace*{\algorithmicindent} \textit{\hspace{1cm} number\_arches (arches)}          \\
	\hspace*{\algorithmicindent} \textit{\hspace{1cm} End-to-end parameters (pars)}     \\
	\hspace*{\algorithmicindent} \textbf{Output:}     				                    \\
	\hspace*{\algorithmicindent} \textit{\hspace{1cm}Multiple objectives  }             \\
	\label{alg:endtoend}
	\begin{algorithmic}[1]
		\Procedure{$end\_to\_end\_eval$}{$ space, arches, pars $}     
		\\
		\LineComment{read input dataset}
		\State $ data \leftarrow read\_database(sub(pars)) $
		\\
		\LineComment{pre-processing the data}
		\State $ data \leftarrow pre\_processing(data, sub(pars)) $
		\\
		\LineComment{compile and train the architecture} 
		\State {$ rarch \leftarrow compile\_train(space, arches, sub(pars)) $}
		\\
		\LineComment{objectives calculation} 
		\State {$objectives \leftarrow obj\_evaluation(rarch)$}
		\\
		\LineComment{return number of potential arches} 
		\State $ return \hspace{0.5cm} objectives $
		
		\EndProcedure
		
	\end{algorithmic}
\end{algorithm}

\textbf{Self-optimization:}
The proposed algorithm is based on the operation of multi-objective Bayesian optimization (MOBO) for NAS, where the novel surrogate and acquisition functions are the most important part of the algorithm. Each function contributes to the improvement and convergence of the algorithm at each loop based on the existing historical runs in the database $H$. 

Generally, given a neural network architecture, the surrogate function can predict in advance the possible outcomes of the architecture. The prediction is used to estimate how promising the architecture is compared with others or with the current trade-off architecture set $P$. A set of $arches$ of promising candidates is selected by the acquisition function. These candidates will be evaluated and updated to the database $H$. Consequently, the newly updated trade-off architectures can be calculated and saved in $P$. 

\begin{algorithm}[t!]
	\caption{An end-to-end self optimization algorithm}
	\hspace*{\algorithmicindent} \textbf{Input:}                         \\
	\hspace*{\algorithmicindent} \textit{\hspace{1cm}1. Initial Neural Architecture Space (inas)}      	  \\
	\hspace*{\algorithmicindent} \textit{\hspace{1cm}2. Heuristics for self-optimization} (heurs)      	  \\
	\hspace*{\algorithmicindent} \textit{\hspace{1cm}3. End-to-end parameters for the pipeline} (pars)    \\
	\hspace*{\algorithmicindent} \textit{\hspace{1cm}4. Termination conditions (iterate) (conds)} \\
	\hspace*{\algorithmicindent} \textbf{Output:}     				     \\
	\hspace*{\algorithmicindent} \textit{\hspace{1cm}Pareto-front P}     \\
	\label{alg:selfoptimization}
	\begin{algorithmic}[1]
		\Procedure{SelfOptAlgs}{$inas,heurs,pars,conds$}   
		
		\While{($ heur \in heurs $)}  \Comment{self-optimized process} 
		\LineComment{1. Initialize potential architectures} 
		\State $space \leftarrow arch\_generation(heur)$
		\State $arches \leftarrow get\_num\_arches(pars)$
		\\
		\LineComment{2. Evaluation of the arches \& store in H}
		\State{$ H \leftarrow end\_to\_end\_eval(space, arches, pars)$}
		\\
		\LineComment{3. Trade-off solutions  are stored in P}
		\State{$ P \leftarrow  \mathit{trade\_off\_calculation} (H) $}
		\\
		\LineComment{4. optimize the arch selection}
		\While {($terminate \leq  conds) $}         
		\LineComment{training surrogate models using H}
		\State $sur \leftarrow surrogate\_model(H)$ 
		
		\LineComment{optimize arch selection}        
		\State $lt\_arcs = acq\_func(sur, space, arches)$
		
		\LineComment{update data for H}
		\State $ H \leftarrow end\_to\_end\_eval(space, l\_arcs, pars)$
		\LineComment{update solutions for P}
		\State $ P \leftarrow update\_P(l\_arcs)$
		
		\LineComment{update terminated condition}
		\State $ terminate \leftarrow terminate + 1$
		\EndWhile
		\\
		\LineComment{self-update heuristic}
		\If{( $ {satisfied\_ solutions} \in P$ )} 
		\State $ break $		\Comment{process ends}
		\Else 
		\State $ heuristic\_selection(heurs)$ 		\Comment{continue}
		\EndIf
		
		\EndWhile 
		\EndProcedure
		
	\end{algorithmic}
\end{algorithm}

\begin{table*}
\label{tab:endtoend-table}
\caption{End-to-end influence on model training}
\vskip 0.1in
\begin{center}
\begin{small}
\begin{sc}
\begin{tabular}{ccccccccccccc}
\toprule
\multicolumn{1}{l}{\multirow{2}{*}{Dataset}} & \multicolumn{4}{c}{\multirow{2}{*}{Filter/Layer}} & \multicolumn{4}{l}{\multirow{2}{*}{Block/Layer}} & \multicolumn{2}{c}{Aug} & \multicolumn{2}{c}{NoAug} \\
\multicolumn{1}{l}{}                         & \multicolumn{4}{c}{}                              & \multicolumn{4}{l}{}                             & SGD        & ADAM       & SGD         & ADAM        \\
\midrule
\multicolumn{1}{l}{\multirow{6}{*}{MNIST}}   & 16         & 16         & 16         & 16         & 1          & 1          & 1          & 2         & 39.36      & 57.37      & 98.96       & 99.02       \\
\multicolumn{1}{l}{}                         & 16         & 16         & 16         & 16         & 1          & 1          & 1          & 1         & 36.93      & 62.00      & 98.79       & 99.23       \\
\multicolumn{1}{l}{}                         & 8          & 16         & 8          & 16         & 1          & 1          & 1          & 2         & 48.22      & 64.48      & 98.57       & 98.80       \\
\multicolumn{1}{l}{}                         & 8          & 16         & 8          & 16         & 1          & 1          & 1          & 1         & 57.75      & 71.61      & 98.65       & 98.95       \\
\multicolumn{1}{l}{}                         & 8          & 8          & 16         & 16         & 1          & 1          & 1          & 2         & 18.26      & 74.99      & 98.64       & 98.98       \\
\multicolumn{1}{l}{}                         & 8          & 8          & 16         & 16         & 1          & 1          & 1          & 1         & 36.50      & 24.04      & 98.41       & 99.08       \\
\midrule
\multirow{6}{*}{Fashion}                     & 64         & 64         & 64         & 64         & 1          & 1          & 1          & 2         & 19.50      & 28.64      & 90.67       & 89.21       \\
                                             & 64         & 64         & 64         & 64         & 1          & 1          & 1          & 1         & 10.04      & 33.86      & 88.71       & 90.97       \\
                                             & 32         & 32         & 32         & 32         & 1          & 1          & 1          & 2         & 13.76      & 12.19      & 89.91       & 91.45       \\
                                             & 32         & 32         & 32         & 32         & 1          & 1          & 1          & 1         & 17.07      & 26.28      & 89.58       & 91.61       \\
                                             & 16         & 16         & 16         & 16         & 1          & 1          & 1          & 2         & 14.14      & 26.26      & 88.41       & 90.01       \\
                                             & 16         & 16         & 16         & 16         & 1          & 1          & 1          & 1         & 21.07      & 62.48      & 89.11       & 91.04       \\
\midrule
\multirow{6}{*}{Cifar10}                     & 128        & 128        & 128        & 128        & 1          & 1          & 1          & 2         & 12.39      & 12.32      & 63.33       & 79.31       \\
                                             & 128        & 128        & 128        & 128        & 1          & 1          & 1          & 1         & 12.92      & 15.94      & 63.35       & 76.92       \\
                                             & 32         & 64         & 128        & 128        & 1          & 1          & 1          & 2         & 23.00      & 24.33      & 56.52       & 80.13       \\
                                             & 32         & 64         & 128        & 128        & 1          & 1          & 1          & 1         & 24.56      & 12.14      & 74.61       & 78.65       \\
                                             & 64         & 128        & 256        & 256        & 1          & 1          & 1          & 2         & 23.08      & 23.99      & 62.53       & 79.48       \\
                                             & 64         & 128        & 256        & 256        & 1          & 1          & 1          & 1         & 18.04      & 28.92      & 76.26       & 74.17       \\
\midrule
\end{tabular}
\end{sc}
\end{small}
\end{center}
\vskip -0.1in
\end{table*}

The basic structure of our algorithm~\ref{alg:selfoptimization} revolves around the MOBO with novel initial space and heuristic. Overall, the algorithm creates an initial neural network architecture space, searching for an optimal architecture, update the surrogate model, and calculate the trade-off results. It is necessary to have a potential initial search space such as mentioned in Section~\ref{InitialSpace}. Following that, the actual end-to-end evaluations of the architectures will be done and stored in the historical database $H$. Next, the trade-off solutions are calculated and saved in $P$. 

The optimization loop begins by training the surrogate models upon the data in $H$, line 8 in the algorithm~\ref{alg:selfoptimization}. Each objective has its model and the model will be used to estimate in advance the outcome for that objective. The detailed operation of this step is described in algorithm~\ref{alg:surrogate}. Because the surrogate model helps to rank the considered neural network architectures, the optimal selection for the ranked architectures simply sorts the rank and takes $arches$ best ones. This step is presented in the algorithm~\ref{alg:acquisition}. Next, the selected $arches$ architectures will be evaluated with actual end-to-end executions by algorithm 3. Finally, the results are used to update both historical data $H$ and trade-off solutions set $P$, lines 20 and 22 respectively.

After the Bayesian optimization loop (the inner while loop in the algorithm~\ref{alg:selfoptimization} completes, the self-optimized loop continues its work. If the trade-off solution set $P$ contains the expected solutions, the loop finishes, otherwise, the loop continues with other heuristics. The way the loop works is similar to the study of humans. The more knowledge or heuristic it has, the better the convergence, and the better solutions might be found, thus, the name of the algorithm is self-optimization. 

\section{Evaluation}
\label{evaluation}

\subsection{Experimental Setup}
In this work, we evaluate our multi-objective end-to-end self-optimized approach using image classification applications and convolution neural network~\cite{CNN:2012}. We use the well-known state-of-the-art Bayesian approach Autokeras~\cite{Auto-keras} as a baseline for evaluating our experiments. Notably, Autokeras supports also manually setting for training a deep neural network, however, we use the default parameters of the tool for evaluating only the automatic part. In addition, We also use the same datasets as state-of-the-art related work for testing.

We aim to have the diversity of NAS that offers different options for users to select based on their demands. Furthermore, the end-to-end evaluation of a neural network has considerable influence on the training results that users should know for a better decision. Notably, our approach reduces the search space, however, it still offers various options to users. This feature is important since a less complicated deep neural network might have competitive accuracy compared with a much more complicated one~\cite{ResNet:2016, WideResNet:2016}. 

In general, we use the state-of-the-art data sets for testing: MNIST, FASHION, and CIFAR10~\cite{Auto-keras}. The experiments are executed on our CSC cloud using a GPU instance that has a single GPU Tesla P100. Each experiment runs from a few minutes to more than 2 days (maximum 3093 minutes) depending on the dataset and the hyper-parameters. Notably, Autokeras~\cite{Auto-keras} does not support the API to set the limited execution time for the tool anymore, therefore, we set the number of models instead. The more models used in Autokeras, the more time it takes to execute.

\subsection{End-to-end Evaluation}
\label{sec:evaluationETE}
In this section, we demonstrate the importance of the end-to-end evaluation of convolution neural networks on image classification by using the augmentation feature of images. Image augmentation usually improves the accuracy of a neural network model. For example, a horizontal flip of a picture of a cat may help for training, because the picture could have been taken from the left or right. However, it is not always the case since it also depends on other features, for instance, training function, learning rate, etc. We demonstrate the results of different neural network architectures with the following cases:
\begin{itemize}
	\item Augmentation: training function Adam $\&$ SGD.
	\item Non-Augmentation: training function Adam $\&$ SGD.
\end{itemize}

We evaluate two cases of the pre-processing stage: one uses the augmentation for the input training images and the other one does not use augmentation. For the sake of illustrating the idea, we use two popular training functions: SGD~\cite{SGD:1991} and ADAM~\cite{Adam:2014} with their default setting parameters. Finally, we run the experiments with different neural network architectures. Concretely, the architectures have various size of the network dimensions: such as the number of filters, number of blocks per layer, and different number of layers. 

\begin{table}[t]
\caption{Autokeras: Accuracy (\%) and Performance (minute)}
\label{sample-table}
\vskip 0.15in
\begin{center}
\begin{small}
\begin{sc}
\begin{tabular}{lcccr}
\toprule
Models\\(Epochs) & Dataset & Accuracy & Performance\\
\midrule
1 (100)& MNIST    & 99.14        & 18.06     \\
2  (20)& $\_$     & 99.19        & 77.04     \\
4  (20)& $\_$     & 99.14        & 1917.22   \\
6  (20)& $\_$     & 99.33        & 2816.85   \\
8  (20)& $\_$     & 98.91        & 3552.54   \\
\midrule
1 (100)& Fashion   & 92.30         & 20.2     \\
2  (20)& $\_$      & 92.25         & 94.84    \\
4  (20)& $\_$      & 94.40         & 1927.06  \\
6  (20)& $\_$      & 91.49         & 2684.45  \\
8  (20)& $\_$      & 94.37         & 3060.48  \\
\midrule
1  (100)& Cifar10   & 72.45        & 18.56   \\
2   (20)& $\_$      & 81.27        & 162.62  \\
4   (20)& $\_$      & 97.91        & 1484.73 \\
6   (20)& $\_$      & 97.68        & 2127.51 \\
8   (20)& $\_$      & 97.68        & 3093.97 \\

\bottomrule
\end{tabular}
\end{sc}
\end{small}
\end{center}
\vskip -0.1in
\end{table}

\begin{table}
\caption{MEESO: Accuracy (\%) and Performance (minute)}
\label{esso-table}
\vskip 0.15in
\begin{center}
\begin{small}
\begin{sc}
\begin{tabular}{lcccr}
\toprule
Arches \\ Blocks & Dataset & Accuracy & Performance\\
\midrule
1 1 1 1 & MNIST    & 99.47        & 8.59        \\
1 1 1 0 & $\_$     & 99.37        & 3.94        \\
2 1 1 1 & $\_$     & 99.37        & 16.44       \\
1 2 1 2 & $\_$     & 99.36        & 5.07        \\
1 1 2 2 & $\_$     & 99.29        & 5.08        \\
\midrule
1 1 1 1 & Fashion   & 92.42         & 21.86     \\
1 1 2 2 & $\_$      & 92.25         & 37.16     \\
1 2 2 1 & $\_$      & 92.40         & 97.48     \\
1 2 1 2 & $\_$      & 92.23         & 26.43     \\
2 2 2 2 & $\_$      & 92.00         & 289.20    \\
\midrule
1 1 1 0 & Cifar10   & 82.92        & 20.36      \\
1 1 1 1 & $\_$      & 81.62        & 54.31      \\
3 4 6 3 & $\_$      & 82.42        & 254.66     \\
3 4 4 3 & $\_$      & 82.01        & 225.13     \\
4 3 5 3 & $\_$      & 81.52        & 149.11     \\

\bottomrule
\end{tabular}
\end{sc}
\end{small}
\end{center}
\vskip -0.1in
\end{table}

\subsection{Baseline Evaluation}
\label{sec:baseline}
This section introduces an evaluation of our (MEESO) and the default automatic features of Autokeras. In Autokeras, there are two default properties: the number of the epoch, and the number of the trial. The former is the number of times to train the dataset and the latter is the number of models to try in each experiment\footnote{$https://autokeras.com/tutorial/image\_classification$}. We evaluate the automatic part of Autokeras in two cases: 
\begin{itemize}
	\item 1 model with different values of epoch from 20 to 100.
	\item 2 to 8 models with the fixed value of epoch 20.
\end{itemize}

We select the range for the epoch to evaluate the tool in short training time, and the range for the model to evaluate its ability to improve accuracy for longer training time. The larger the number of models, the longer the execution time is. We set the maximum number of the model to 8 for limiting the experimental time to the maximum of three days and the experiments run on only one GPU P100.

We randomly initialize a set of architectures constructed upon the ResNet itentity block~\cite{ResNet:2016} and additional Dropout~\cite{Dropout-Bayesian-2016} as described in section~\ref{sec:end-to-end-optimization}. We run our self-optimized algorithm for the same amount of time and extract $K$ best optimal solutions based on only accuracy metric for a fair evaluation. Notably, the historical data are also generated during this time (see algorithm~\ref{alg:selfoptimization}).

\subsection{Multi-objective Evaluation}
\label{sec:multiobjecitiveEva}
Deep learning models are rarely driven by a single objective, and most often, require trade-offs for many competing objectives~\cite{NSGA-Net:2019, NSGA-Net:2020}. We estimate the uncertainty of model prediction that allows users to estimate the new data. To evaluate the uncertainty of neural network architecture, we generate random input images which are completely different from the dataset and calculate the uncertainty following the formulations presented in Section~\ref{considerobjs}. 

In this experiment, we increase the number of training time (epoch) ranging from $100$ to $1000$. The aim of this is to evaluate the self-optimization feature of the algorithm when more time is available. Since the search space for end-to-end approach is huge, we only use the number of training time in this experiment, however, the extension of this experiment with other hyper-parameters are also useful with the self-optimized approach. We leave this to future work since it requires longer experiments. 

The goal of this multi-objective evaluation is to offer various trade-off solutions for users to select upon their demands.

\begin{figure*}[h]
	\centering
	\includegraphics[scale=0.75]{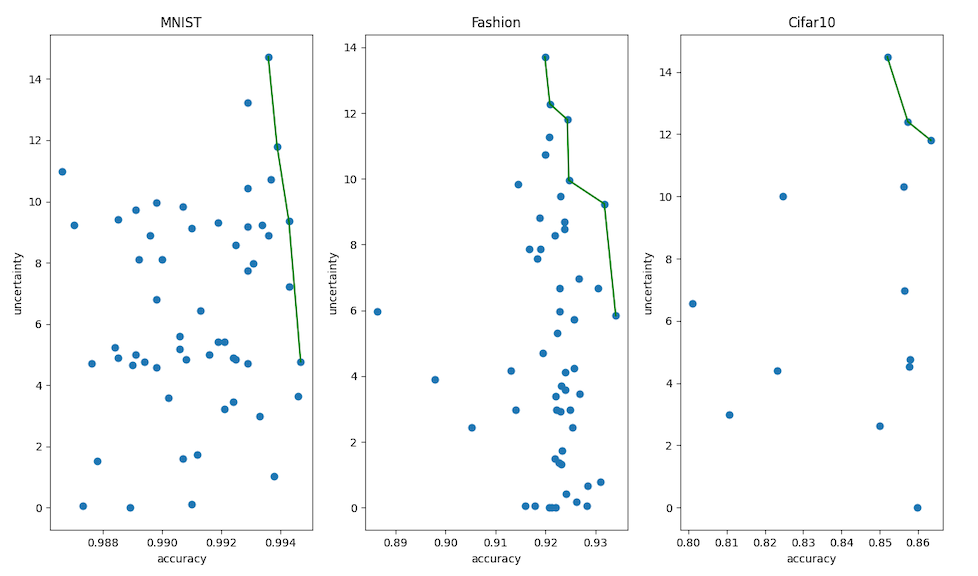}
	\caption{Trade-off solutions for two metrics: accuracy and uncertainty}		\label{fig:tradeoff}
\end{figure*}
	
\section{Results Analysis}
\label{resultsanalysis}
\subsection{End-to-End Influence}
For this experiment, we analyze the obtained accuracy of different neural network architectures as being illustrated in Table 1. The Table presents 6 typical different architectures corresponding with each dataset following our experiments. Each architecture has a different number of filters and blocks in each layer. The results show that Augmentation does not help much for the considered datasets. Conversely, the Non-Augmentation case usually has much better accuracy regardless of training functions. In addition, the training function Adam usually gives better accuracy than SGD, however, both functions have competitive accuracy in the Non-Augmentation case.

The results confirm the importance of end-to-end evaluation. Data pre-processing is one of the most important steps that help to improve the performance and accuracy~\cite{Hutter2019AutomatedML}. The input data usually requires to be pre-processed or concretely augmentation in the case of image data, before giving to the model for training. In some situations, data pre-processing is not necessary since there is no improvement in accuracy to be observed. Nevertheless, when the selected model cannot provide expected results, the pre-processing step is usually a useful step to improve the prediction accuracy. Clearly, the other features can also affect the training results such as hyper-parameters. For the sake of illustration, we do not present all the effects of the hyper-parameters in this section. However, they play an important role in increasing the accuracy of the model~\cite{AutoML:2019}, therefore, an end-to-end evaluation is always useful for AutoDL.

\subsection{Baseline Analysis}

This experiment is described in section \ref{sec:evaluationETE} and its results are presented in Tables $2$ and $3$ for both Autokeras and MEESO algorithm respectively. In this experiment, we select $K=5$ for extracting the five optimal architectures. To briefly present the results, we present only the number of blocks in each layer that typically describe the neural network architectures (see Table 3).

We obtain various architectures that have competitive accuracy compared with Autokeras in a short training time. For longer training time, Autokeras gives better accuracy. The main reason for the obtained results is the Autokeras fine-tunes the model, while our MEESO targets to discover various architectures as well as support multiple constraints. The results obtained for the MNIST dataset illustrate the importance of the AutoDL diversity, where the smaller models are preferred because they give the competitive accuracy with much less performance. Furthermore, training such simple models do not require much computing resources.

This experiment also shows that MEESO can discover different potential architectures that empower users to choose upon their demands. However, this is a trade-off for AutoDL, where the end-to-end training requires more time to achieve better accuracy. The results also show that an AutoDL approach should consist of two steps: discover potential architectures and fine-tune those architectures. In addition, this introduces an challenge for the diverse approach of AutoDL: after a set of potential architectures are discovered, it still requires substantial effort to achieve the desired accuracy. 


\subsection{Multi-objective Solutions}
\label{subsec:Multiobjective-solutions}

Building and selecting a DNN models is a multi-objective optimization problem, in which trade-offs
between accuracy and uncertainty are considered in this work. Figure 1 gives an overview of the trade-off solutions obtained from this experiment. We have observed that the more time the algorithm has, the better accuracy is obtained. Both Fashion and Cifar10 achieve better results in this experiment compared with the results of the Table 3 in section \ref{sec:baseline}. 

The trade-off solutions or the current Pareto front in Figure 1 can support user to make better decisions. For simple dataset such as MNIST and FASHION, more trade-off solutions are discovered compared with the CIFAR10 in the limit of experimental time. Users may be interested in choosing a model that has a balance between accuracy and uncertainty metrics, so they can identify the novel input data. In other words, users can be based on the measurement of the uncertainty metric to decide how certain the prediction is, so they can decide whether to recheck the input data if the uncertainty is high. In addition, ensemble learning can be used to combine the predictions from multiple neural network models to reduce generalization error~\cite{ensemble:2021}.



\section{Conclusions and future works}
\label{conclusions}

In this paper, we propose a multi-objective end-to-end self-optimized approach for building deep learning models. Our approach is a heuristic approach for generating the initial search space of neural network architectures. Following this, we introduce a novel surrogate function which reduces the search space into the groups of potential architectures. The surrogate function efficiently evaluates the potential architectures in advance, so the number of actual training executions could be reduced significantly. 

In addition, our approach also keep the diversity of AutoDL by introducing to users various neural network architectures and a multi-objective evaluation in terms of accuracy and uncertainty. The multi-objective AutoDL is necessary since it can effectively and efficiently deal with different type of data sets. Clearly, the less complicated network architecture is preferred if it provides the same accuracy, such as the case of MNIST and Fashion since it requires less computing resource and performance. Our empirical experiments also show that the end-to-end approach has an important role in training neural networks. However, it still requires additional effort to have an extension of end-to-end evaluation that fine-tunes the training stage. 

Automatic deep learning is becoming an important tool in modern computing applications. However, AutoDL is a time-consuming tool. This is a big obstacle for users to use AutoDL for their data. As a result, a parallelization of the AutoDL tool which accelerates the training of multiple neural networks on a distributed system is necessary. Parallel and distributed training of a DNN has been widely considered~\cite{Hoefler:2019,Mayer:2020} by related works, however, the parallelization of multiple neural networks have not been investigated so far. We leave this problem to our future work since it requires additional novel contributions. 


\section*{Acknowledgment}
This work has been conducted mostly at Aalto University since 2020. The author would like to thank Assoc. Prof. L. Truong for his support and Prof. K. Heljanko for some discussions.

 \bibliographystyle{elsarticle-num} 
 \bibliography{cas-refs}










\end{document}